\documentclass[lettersize,journal]{IEEEtran}

\usepackage{amsmath,amsfonts,amssymb}
\usepackage{graphicx}
\usepackage{booktabs}
\usepackage{makecell}
\usepackage{threeparttable}
\usepackage{multirow}
\usepackage[hidelinks]{hyperref}

\usepackage[
    backend=biber,
    style=ieee,
    sorting=none
]{biblatex}
\begin{document}

\title{Value-Decomposed Reinforcement Learning Framework for Taxiway Routing with Hierarchical Conflict-Aware Observations}

\author{Shizhong~Zhou, Haifeng~Liu, Zheng~Zhang, Shiyu~Zhang, Bo~Yang, and~Yi~Lin%
\thanks{This work has been submitted to the IEEE for possible publication. Copyright may be transferred without notice, after which this version may no longer be accessible.}%
\thanks{Corresponding author: Yi Lin; e-mail: yilin@scu.edu.cn}%
\thanks{Shizhong Zhou is with the National Key Laboratory of Fundamental Science on Synthetic Vision, Sichuan University, Chengdu 610000, China. Haifeng Liu, Zheng Zhang, Shiyu Zhang, Bo Yang, and Yi Lin are with the College of Computer Science, Sichuan University, Chengdu 610000, China.}}

\maketitle

\begin{abstract}
Taxiway routing and on-surface conflict avoidance are coupled safety-critical decision problems in airport surface operations. Existing planning and optimization methods are often limited by online computational cost, while reinforcement learning methods may struggle to represent downstream traffic conflicts and balance multiple objectives. This paper presents Conflict-aware Taxiway Routing (CaTR), a reinforcement learning framework for real-time multi-aircraft taxiway routing. CaTR constructs a grid-based airport surface environment with action masking, introduces a hierarchical foresight traffic representation to encode current and downstream conflict-related traffic conditions, and adopts a value-decomposed reinforcement learning strategy to prioritize sparse but safety-critical objectives. Experiments are conducted on a realistic environment based on Changsha Huanghua International Airport under multiple traffic density levels. Results show that CaTR achieves better safety--efficiency trade-offs than representative planning, optimization, and reinforcement learning baselines while maintaining practical runtime.
\end{abstract}

\begin{IEEEkeywords}
Taxiway Routing, Reinforcement Learning, Adaptive Value Weighting, Conflict-Aware Routing, Discrete Grid Topology, Multi-Objective Optimization
\end{IEEEkeywords}

\section{Introduction}
\IEEEPARstart{A}{s} air traffic demand continues to increase, taxiway routing and on-surface conflict avoidance have become important factors affecting airport capacity, taxi delay, fuel consumption, and operational safety. During dense airport surface operations, multiple aircraft compete for limited taxiway, runway crossing, holding, and gate access resources. A local routing decision may affect downstream segment occupancy and further propagate congestion or conflict risks through the airport surface network.

Airport taxiway routing has been studied using planning and optimization methods, including mixed-integer programming, chance-constrained optimization, path search, and metaheuristic algorithms~\cite{clare2011optimization,yang2022stochastic,wang2021chance,zhang2024research,deng2022multi}. These methods can provide interpretable routing plans under explicit constraints, but their online use remains challenging when traffic conditions change rapidly. Repeated optimization may become computationally expensive, and real-time congestion or conflict information may be difficult to incorporate effectively.

Reinforcement learning (RL) offers a promising alternative by learning policies that map traffic states to routing actions in dynamic environments. Recent RL-based studies have been applied to airport taxiing, departure metering, and multi-aircraft coordination~\cite{xiang2023application,szymanski2023single,ali2022deep,watteau2024optimizing,tran2024towards}. However, existing RL methods still face three challenges in dense taxiway routing. First, local observations may not sufficiently capture downstream congestion and conflict propagation. Second, different operational objectives, such as safety and routing efficiency, may produce conflicting optimization signals. Third, safety-critical events are sparse during training, which may weaken policy robustness in high-density scenarios.

To address these limitations, this paper proposes Conflict-aware Taxiway Routing (CaTR), an RL-based framework for real-time airport surface route decision making. CaTR combines a grid-based discrete airport surface environment, a hierarchical foresight traffic representation (HFTR), and a value-decomposed learning strategy. The main contributions are summarized as follows:
\begin{itemize}
    \item A grid-based airport surface routing framework is constructed with action masks that encode taxiway connectivity, cell occupancy, and runway operation constraints.
    \item A hierarchical foresight traffic representation is employed to describe current and downstream traffic conditions along candidate taxiway segments for conflict-aware decision making.
    \item A value-decomposed reinforcement learning strategy is introduced to balance multiple reward components and prioritize sparse but safety-critical objectives.
    \item Experiments on a realistic Changsha Huanghua International Airport environment demonstrate that CaTR outperforms representative baselines in safety--efficiency trade-offs across different traffic densities.
\end{itemize}

\section{Problem Formulation}
\subsection{Grid-Based Airport Surface Environment}

\begin{figure}[t]
  \centering
  \includegraphics[width=0.95\linewidth]{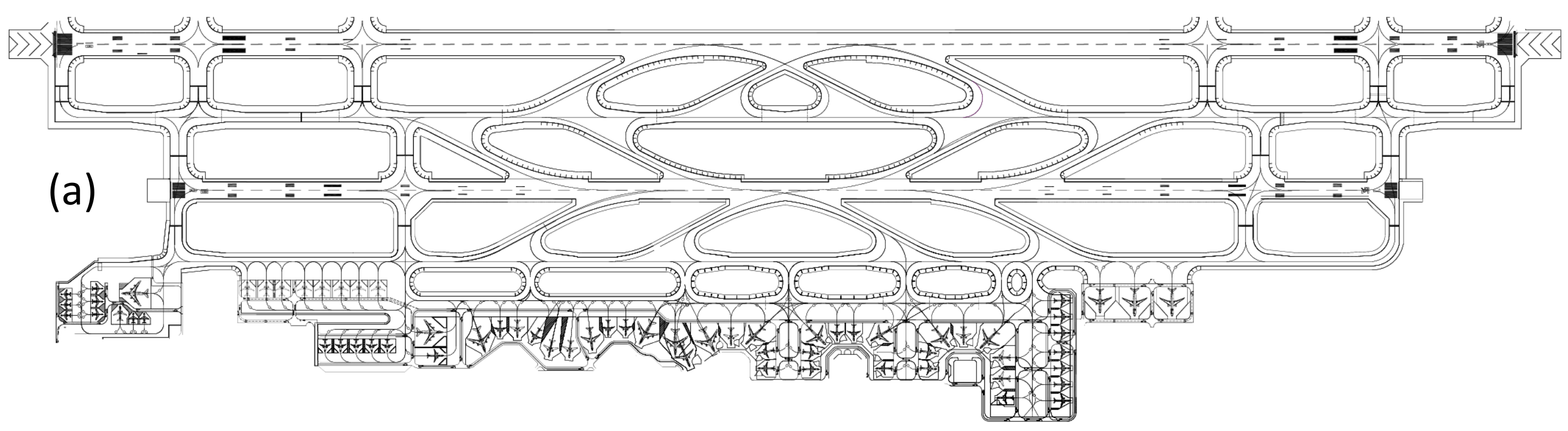}\\[1ex]
  \includegraphics[width=0.95\linewidth]{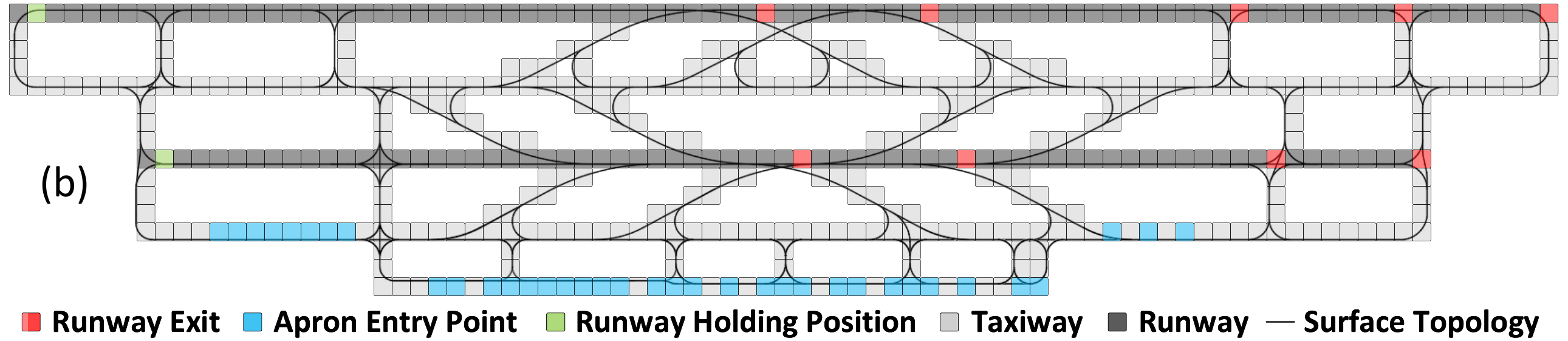}
  \caption{(a) Real-world airport taxiway--runway layout. (b) Corresponding grid-based discretization.}
  \label{fig:grid_model}
\end{figure}

As shown in Fig.~\ref{fig:grid_model}, the airport surface is discretized into uniform grid cells to model taxiway and runway operations. Each grid state is represented by
\begin{equation}
\mathbf{c}=(x,y,h),
\end{equation}
where $x$ and $y$ denote the grid coordinates and $h\in\{0,1,2,3\}$ denotes the discretized aircraft heading. The discrete action set is
\begin{equation}
\mathcal{A}=\{a^{\mathrm{forward}},a^{\mathrm{stop}},a^{\mathrm{left}},a^{\mathrm{right}}\}.
\end{equation}
At each decision step, an action mask excludes infeasible actions according to taxiway topology, heading constraints, grid-cell occupancy, and runway operation rules. This ensures that aircraft movements remain physically feasible and comply with basic surface operation constraints.

For each aircraft, the state at time step $t$ consists of a routing state and a conflict-related traffic observation:
\begin{equation}
s_t=(s_t^{\mathrm{route}},s_t^{\mathrm{HFTR}}).
\end{equation}
The routing state is defined as
\begin{equation}
s_t^{\mathrm{route}}=
(x_t,y_t,h_t,\delta x_t,\delta y_t,\eta_t^{(1)},\tau_t^{(1)},\eta_t^{(2)},\tau_t^{(2)}),
\end{equation}
where $(x_t,y_t,h_t)$ denotes the current grid position and heading, $(\delta x_t,\delta y_t)$ denotes the relative offset to the destination, $\eta_t^{(k)}$ denotes the state of runway $k$, and $\tau_t^{(k)}$ denotes the remaining time steps before runway $k$ becomes available. The runway state includes empty, takeoff, landing, and crossing modes.

\subsection{Learning Objective}
The taxiway routing problem is formulated as a Markov decision process. The policy selects valid actions under the action mask:
\begin{equation}
a_t \sim \pi_\theta(\cdot\mid s_t,\mathbf{m}_t), \qquad a_t\in\mathcal{A}_t^{\mathrm{valid}}.
\end{equation}
The objective is to maximize the expected discounted return:
\begin{equation}
J(\theta)=\mathbb{E}_{\pi_\theta}\left[\sum_{t=0}^{\infty}\gamma^t r_t\right].
\end{equation}
In this work, PPO with action masking is used for policy optimization.

\section{Method}
\subsection{Overview of CaTR}
CaTR is designed to support real-time conflict-aware taxiway routing in dense airport surface operations. The framework consists of three main components: hierarchical foresight traffic representation, dual-branch feature extraction, and gradient-weighted value decomposition. All aircraft share the same policy and value networks, enabling scalable decision making under changing traffic conditions.

\begin{figure*}[t]
  \centering
  \includegraphics[width=0.95\textwidth]{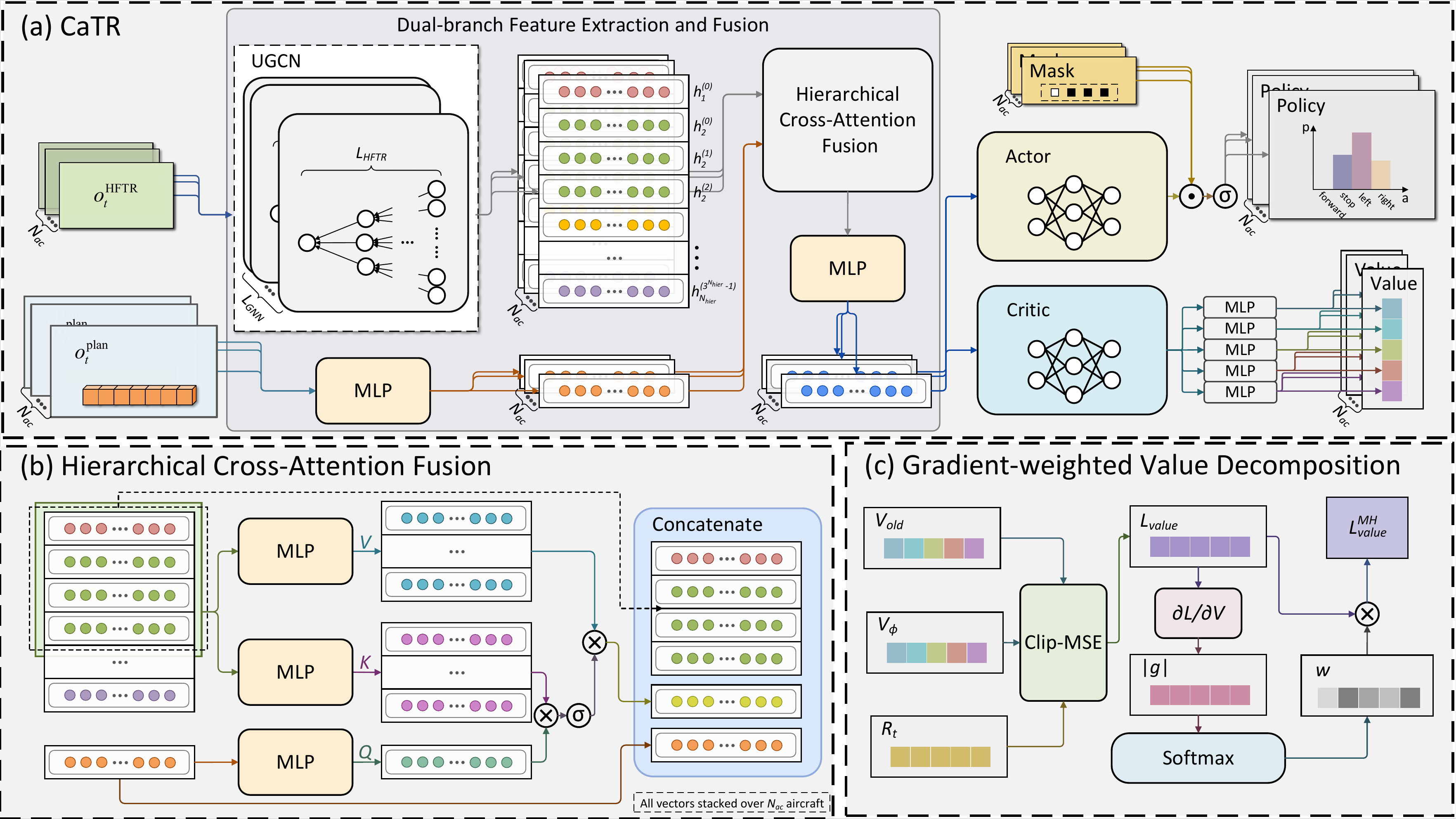}
  \caption{Overall architecture of the proposed CaTR framework.}
  \label{fig:framework}
\end{figure*}

\subsection{Hierarchical Foresight Traffic Representation}
The hierarchical foresight traffic representation (HFTR) is used to encode conflict-related traffic information around the controlled aircraft. Starting from the current taxiway segment, HFTR recursively expands along feasible downstream maneuver options, including left turn, straight ahead, and right turn. This provides a structured description of anticipated traffic situations and potential conflict risks along future taxi paths.

For each taxiway segment, a compact feature vector is defined as
\begin{equation}
\tau_i=[\mathrm{ID}(\tau_i),l_i^{\mathrm{rem}},d_i^{\mathrm{head}},d_i^{\mathrm{same}},n_i],
\end{equation}
where $\mathrm{ID}(\tau_i)$ denotes the segment identifier, $l_i^{\mathrm{rem}}$ denotes the remaining segment length, $d_i^{\mathrm{head}}$ and $d_i^{\mathrm{same}}$ denote the nearest head-on and same-direction aircraft distances, and $n_i$ denotes the number of aircraft on the segment. The HFTR at time $t$ is written as
\begin{equation}
s_t^{\mathrm{HFTR}}=\{\mathcal{T}^{(0)},\mathcal{T}^{(1)},\ldots,\mathcal{T}^{(L_{\mathrm{HFTR}}-1)}\},
\end{equation}
where $\mathcal{T}^{(l)}$ is the set of segment features at the $l$-th foresight level.

\subsection{Dual-Branch Feature Extraction}
The routing state and HFTR describe complementary information. The routing state provides self-related navigation information, while HFTR provides structured downstream traffic information. Therefore, CaTR adopts a dual-branch feature extraction module. The routing branch encodes $s_t^{\mathrm{route}}$ using multilayer perceptrons, and the traffic branch encodes $s_t^{\mathrm{HFTR}}$ using graph-based feature extraction over the hierarchical foresight structure. The extracted features are fused and then provided to the actor--critic networks for masked action selection and value estimation.

\subsection{Gradient-Weighted Value Decomposition}
Airport surface routing involves multiple objectives, including destination completion, route efficiency, separation maintenance, and conflict avoidance. The reward is decomposed as
\begin{equation}
r_t=r_t^{\mathrm{dist}}+r_t^{\mathrm{move}}+r_t^{\mathrm{arrive}}+r_t^{\mathrm{prox}}+r_t^{\mathrm{conf}},
\end{equation}
where the components correspond to distance, movement, arrival, proximity, and head-on conflict objectives, respectively.

To balance these objectives, the critic estimates a value vector for different reward components. For the $i$-th value component, the gradient magnitude of the value loss is computed as
\begin{equation}
g_i=\left\|\nabla_{\phi}\mathcal{L}_{\mathrm{value},i}\right\|_1.
\end{equation}
The normalized component weight is then obtained by
\begin{equation}
w_i=\frac{\exp(g_i)}{\sum_{j=1}^{K}\exp(g_j)}.
\end{equation}
The total value loss is defined as
\begin{equation}
\mathcal{L}_{\mathrm{value}}=K\sum_{i=1}^{K}w_i\mathcal{L}_{\mathrm{value},i},
\end{equation}
where $K$ is the number of reward components. This strategy increases the learning emphasis on difficult or sparse objectives, especially safety-critical conflict avoidance.

\section{Experiments}
\subsection{Experimental Setup}
Experiments are conducted in a realistic airport surface environment based on Changsha Huanghua International Airport (ZGHA), whose parallel runways require runway crossing operations for all flights. Flight trajectories collected at ZGHA from June 14 to June 27, 2025 are used to characterize traffic density. The maximum hourly traffic density is 42 aircraft, which is used as the base traffic density. Experiments are conducted at $1.00\times$, $1.25\times$, and $1.50\times$ traffic densities.

CaTR is compared with planning, optimization, and reinforcement learning baselines, including Dijkstra, Improved A*, GA, DQN, and PPO. The evaluation metrics include head-on conflict rate (HCR), proximity conflict rate (PCR), success rate (SR), detour ratio (DR), excess time ratio (ETR), and runtime (RT). Lower HCR, PCR, DR, ETR, and RT indicate better performance, while higher SR indicates better performance.

\begin{table*}[t]
  \footnotesize
  \centering
  \caption{Performance of baseline methods under different traffic densities.}
  \label{tab:baseline-methods-comparison}
  \begin{threeparttable}
    \renewcommand{\arraystretch}{1.12}
    \begin{tabular}{l l c c c c c c}
      \toprule
      \makecell[c]{\textbf{Traffic}\\\textbf{Density}}
        & \makecell[c]{\textbf{Method}}
        & \multicolumn{2}{c}{\makecell[c]{\textbf{Conflict Avoidance}}}
        & \multicolumn{3}{c}{\makecell[c]{\textbf{Routing Efficiency}}}
        & \makecell[c]{\textbf{Computational}\\\textbf{Efficiency}} \\
      \cmidrule(lr){3-4} \cmidrule(lr){5-7} \cmidrule(lr){8-8}
      &
        & \textbf{HCR (\%)} $\downarrow$
        & \textbf{PCR (\%)} $\downarrow$
        & \textbf{SR (\%)} $\uparrow$
        & \textbf{DR (\%)} $\downarrow$
        & \textbf{ETR (\%)} $\downarrow$
        & \textbf{RT (s)} $\downarrow$ \\
      \midrule
      \multirow{6}{*}{$1.00\times$}
        & Dijkstra      & 10.005 & 264.70 & 81.69 &  8.21 & 118.30 & \underline{1.59} \\
        & Improved A*   &  5.923 &  91.95 & 91.47 &  6.91 &  52.42 & 5.72 \\
        & GA            &  4.385 & 105.70 & 91.19 & \underline{1.75} & 51.50 & 307.23 \\
        & DQN           &  2.954 & 163.89 & 85.27 & 15.26 & 145.80 & \textbf{0.94} \\
        & PPO           & \underline{1.882} & \underline{35.94} & \underline{95.46} & 4.85 & \underline{50.80} & 2.65 \\
        & \textbf{CaTR} & \textbf{0.169} & \textbf{3.50} & \textbf{99.47} & \textbf{1.67} & \textbf{21.67} & 3.77 \\
      \midrule
      \multirow{6}{*}{$1.25\times$}
        & Dijkstra      & 10.514 & 374.39 & 77.50 & 11.14 & 162.91 & \underline{2.01} \\
        & Improved A*   &  6.580 & 124.05 & 89.49 &  8.55 &  66.62 & 7.76 \\
        & GA            &  4.000 & 184.00 & 88.90 & \underline{3.33} & 77.78 & 347.36 \\
        & DQN           &  2.871 & 140.19 & 87.69 & 14.82 & 117.55 & \textbf{0.88} \\
        & PPO           & \underline{2.112} & \underline{50.06} & \underline{94.43} & 4.96 & \underline{64.27} & 2.62 \\
        & \textbf{CaTR} & \textbf{0.216} & \textbf{7.91} & \textbf{98.78} & \textbf{2.04} & \textbf{29.37} & 3.51 \\
      \midrule
      \multirow{6}{*}{$1.50\times$}
        & Dijkstra      & 10.056 & 450.50 & 73.63 & 13.83 & 199.11 & \underline{2.68} \\
        & Improved A*   &  6.179 & 160.63 & 86.80 &  9.83 & \underline{81.29} & 3.87 \\
        & GA            &  5.333 & 238.75 & 84.88 & \underline{4.96} & 95.77 & 427.66 \\
        & DQN           &  2.958 & 256.18 & 79.96 & 21.50 & 191.76 & \textbf{0.97} \\
        & PPO           & \underline{1.961} & \underline{74.00} & \underline{92.50} & 6.01 & 83.96 & 2.69 \\
        & \textbf{CaTR} & \textbf{0.228} & \textbf{12.18} & \textbf{96.84} & \textbf{2.73} & \textbf{36.88} & 3.48 \\
      \bottomrule
    \end{tabular}
    \begin{tablenotes}[flushleft]\scriptsize
      \item The best results are shown in bold, and the second-best results are underlined.
    \end{tablenotes}
  \end{threeparttable}
  \normalsize
\end{table*}

\subsection{Comparison with Baselines}
Table~\ref{tab:baseline-methods-comparison} reports the performance of CaTR and the baselines under different traffic densities. CaTR achieves the best overall safety--efficiency trade-off across all evaluated traffic densities. Compared with planning and optimization baselines, CaTR substantially reduces conflict rates while maintaining high routing efficiency. Compared with RL baselines, CaTR provides stronger conflict avoidance and more stable completion performance under dense traffic conditions.

At the highest traffic density, CaTR reduces HCR to $0.228\%$ and PCR to $12.18\%$, while maintaining a success rate of $96.84\%$. These results indicate that the proposed conflict-aware representation and value-decomposed optimization improve safety without relying on excessive detours or delays. Although DQN and Dijkstra have lower runtime, their conflict avoidance and routing efficiency are substantially weaker. CaTR therefore provides a favorable trade-off between operational safety, routing efficiency, and computational cost.

\printbibliography

\end{document}